\documentclass{sig-alternate}
\usepackage[tight,footnotesize]{subfigure}
\usepackage{graphicx}
\usepackage{amsmath}
\usepackage{amssymb} 
\usepackage{color}
\usepackage{algorithm}
\usepackage{algorithmic}
\usepackage{times}
\usepackage{bm}
\usepackage{multirow}
\usepackage{mathrsfs}
\usepackage{threeparttable}
\usepackage{amsmath}
\usepackage{dcolumn}
\usepackage{multirow}
\usepackage{booktabs}
\usepackage{threeparttable}

\begin{document}
%

\title{Seeing the Big Picture:\\ Deep Embedding with Contextual Evidences }

\numberofauthors{4} 
%
\author{
%
%
Liang Zheng$^{1}$, Shengjin Wang$^\text{1}$, Fei He$^\text{1}$, Qi Tian$^\text{2}$\\
       \affaddr{$^\text{1}$Department of Electronic Engineering, Tsinghua University, Beijing, China}\\
       \affaddr{$^\text{2}$University of Texas at San Antonio, TX, USA}\\
       \email{\{zheng-l06, hef05\}@mails.tsinghua.edu.cn, wgsgj@tsinghua.edu.cn, qitian@cs.utsa.edu}
}

\maketitle
\begin{abstract}
In the  Bag-of-Words (BoW) model based image retrieval task, the precision of visual matching plays a critical role in improving retrieval performance. Conventionally, local cues of a keypoint are employed. However, such strategy does not consider the contextual evidences of a keypoint, a problem which would lead to the prevalence of false matches. To address this problem, this paper defines ``true match'' as a pair of keypoints which are similar on three levels, \emph{i.e.,} local, regional, and global. Then, a principled probabilistic framework is established, which is capable of implicitly integrating discriminative cues from all these feature levels.

Specifically, the Convolutional Neural Network (CNN) is employed to extract features from regional and global patches, leading to the so-called ``Deep Embedding'' framework. CNN has been shown to produce excellent performance on a dozen computer vision tasks such as image classification and detection, but few works have been done on BoW based image retrieval. In this paper, firstly we show that proper pre-processing techniques are necessary for effective usage of CNN feature. Then, in the attempt to fit it into our model, a novel indexing structure called ``Deep Indexing'' is introduced, which dramatically reduces memory usage.

Extensive experiments on three benchmark datasets demonstrate that, the proposed Deep Embedding method greatly promotes the retrieval accuracy when CNN feature is integrated. We show that our method is efficient in terms of both memory and time cost, and compares favorably with the state-of-the-art methods.
\end{abstract}

\category{H.4}{Information Systems Applications}{Miscellaneous}
\category{D.2.8}{Software Engineering}{Metrics}[complexity measures, performance measures]

\terms{Theory}

\keywords{Image retrieval, Bag-of-Words, Contextual cues, Deep Embedding}

\section{Introduction}
This paper considers the task of large scale image retrieval.
Our goal is to retrieve in a large database all the similar images with respect to the query. Over the last decade, considerable efforts have been devoted to improving retrieval performance. One milestone was established by the introduction of SIFT \cite{SIFT2} feature. The state-of-the-art methods in image retrieval mostly employ this low-level feature, which forms the basis of the Bag-of-Words (BoW) model.

\begin{figure}
  \centering
  \includegraphics[width=3.3in]{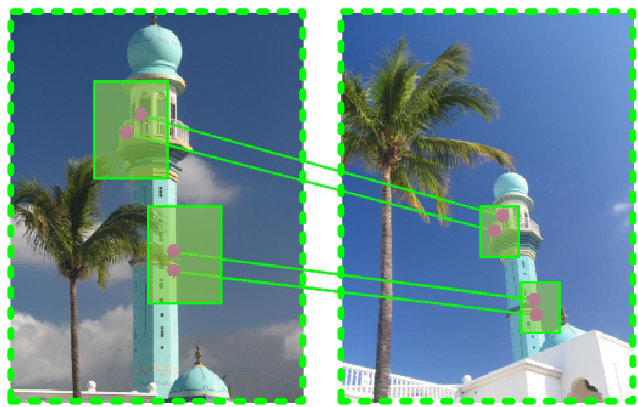}\\
  \caption{An example of true match between keypoints (from the Holidays \cite{Hamming} dataset). In this paper, the true match of a given keypoint is required to be visually similar on three levels, \emph{i.e.,} local, regional, as well as global.}\label{fig:true_match_example}
\end{figure}

Inspired from the well-developed text retrieval routine, the BoW model transforms an image into a histogram of \emph{visual words} through feature quantization. A \emph{codebook}, in which the visual words are defined, is obtained on a pool of SIFT features by unsupervised clustering algorithms \cite{AKM, HKM}. The visual words are weighted by TF-IDF scheme \cite{video_google, zheng2013lp}. To improve efficiency, an inverted file is constructed to perform retrieval in real time.

\begin{figure}
  \centering
  \includegraphics[width=3.3in]{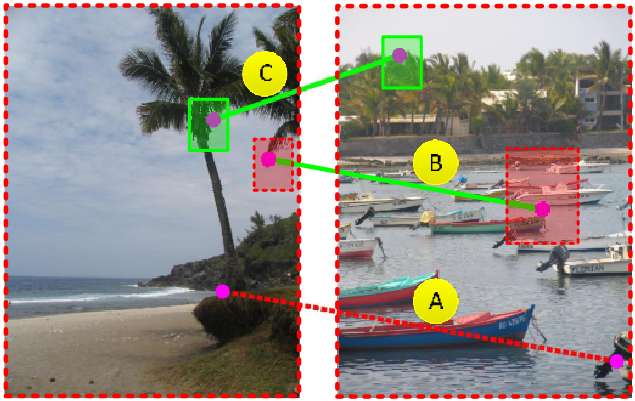}\\
  \caption{Example of false matches. (A): two keypoints are of the same visual word but have a large SIFT Hamming distance. (B): keypoints are similar in SIFT feature, but dissimilar in regional contexts. (C): keypoints are similar in both local and regional features, but belong to irrelevant images (global).  }\label{fig:false_match}
\end{figure}

Basically, visual matching is an essential issue in BoW model. A pair of keypoints are considered as a match if their local features are quantized to the same visual word. But visual word based matching is too coarse and leads to false matches. An effective solution to this problem is to use local cues to determine matching strength. An example of this idea includes Hamming Embedding \cite{Hamming}, which refines this process by computing the Hamming distance between their binary signatures (as in Fig. \ref{fig:false_match}-A). However, one important aspect is neglected: the contextual cues on a larger region around the keypoint are not taken into account. Previous works \cite{BOC, zheng2014coupled} propose to use local color features as a contextual cue. But these methods are generally heuristic for the lack of theoretical interpretation.

To address this problem, this paper proposes to use contextual evidences from multiple levels to improve matching accuracy. Departing from \cite{BOC, zheng2014coupled}, our work employs regional and global contexts. As is shown by Fig. \ref{fig:false_match}-B and Fig. \ref{fig:false_match}-C, contextual evidences can be used to filter out false matches. In this paper, two keypoints are defined as a true match if and only if (\emph{iff}) they are similar on all three feature levels, \emph{i.e.,} local, regional, and global (Fig. \ref{fig:true_match_example}). Starting with this assertion, a probabilistic model is constructed to model the visual matching process. We show that the matching confidence can be implicitly formulated as the product of matching strengths on three levels respectively, thus providing a principled framework on how multi-level features can be combined in the BoW model.

Specifically, to describe regional and global characteristics, the convolutional neural network (CNN) \cite{lecun1998gradient} is employed. This learning machine has been applied in various computer vision tasks \cite{krizhevsky2012imagenet, girshick2013rich} and yielded surprisingly good performance. Nevertheless, few works have been done in the field of image retrieval, especially on how CNN feature can be adapted in the BoW model. To this end, this paper firstly demonstrates the necessity of pre-processing steps for proper usage of CNN feature. Then, regional and global CNN features are are fed into the probabilistic model, yielding the ``Deep Embedding'' framework. We show that, CNN feature is very effective in providing complementary cues to SIFT feature, and that compared with color histogram, it has greater capacity in describing images with large variations.

Overall, this paper claims three major contributions. First, we present a principled probabilistic model for feature fusion at local, regional, and global levels. We show that our model greatly reduces the impact of false matches. Second, the convolutional neural network (CNN) is employed to extract features from image patches, yielding the Deep Embedding framework. Through the introduction of Deep Indexing structure, we provide a possible way to adapt CNN in the BoW model. Finally, state-of-the-art performance is reported on the benchmarks.

The rest of this paper is organized as follows. After a brief review of the related work in Section \ref{section:related_work}, we describe the feature design scheme in Section \ref{section:feature_design}. Then, Section \ref{section:deep_embedding} introduces the Deep Embedding framework in detail. The experimental results are summarized in Section \ref{section:experiments}. Finally, Section \ref{section:conclusions} concludes the paper.

\section{Related Work}
\label{section:related_work}
In the BoW model, due to the ambiguity nature \cite{ambiguity} of visual word, it is important to inject discriminative power to the final representation. One solution to this problem includes modeling spatial constraints \cite{shen2012object, gavves2010landmark, zheng2013visual}: matched features should pass spatial consistency check to filter out false matches. It is also feasible to merge multiple codebooks \cite{zheng2014bayes, jegou2012negative} to compensate for information loss. Another strategy includes feature fusion. Heterogenous features, such as color \cite{zheng2014coupled, BOC}, provide complementary cues in addition to the visual word, enhancing its discriminative power.  Meanwhile, it is effective to preserve binary signatures from the original descriptor. Examples include Hamming Embedding \cite{Hamming}, which computes a Hamming distance between signatures to further verify the matching strength. Similar ideas are also reflected in \cite{qin2013query, selective_match}. This line of approaches are successful because much discriminative power is preserved in the binary features which are very efficient in both memory and matching speed.

 \makeatother
\begin{figure*} [t]
\centering
\subfigure[1st scale]{\label{fig:level_one_partition}%
\includegraphics[width=2.1in]{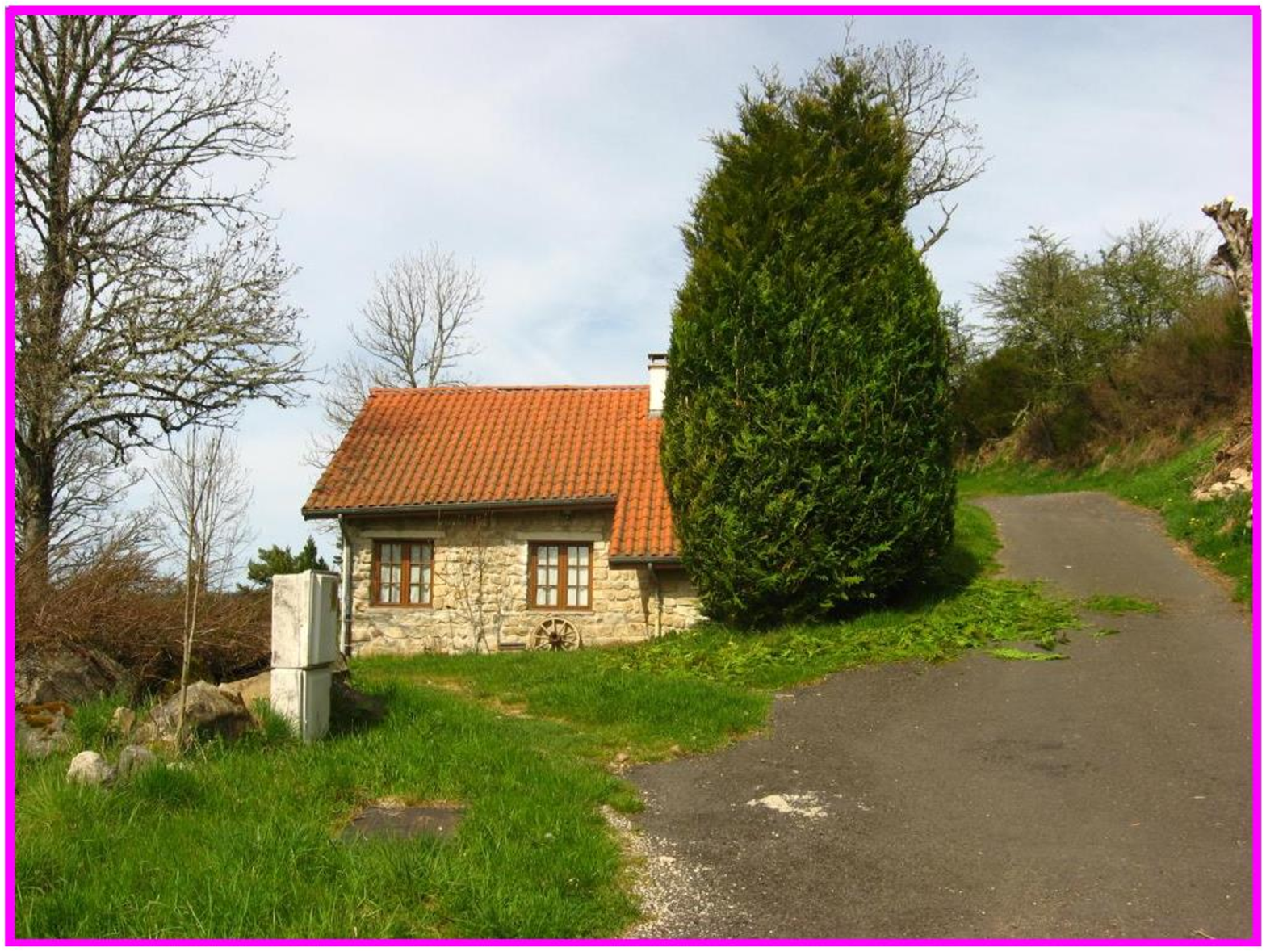}}
\hspace{0.2in}
 \subfigure[2nd scale]{\label{fig:level_two_partition}%
\includegraphics[width=2.1in]{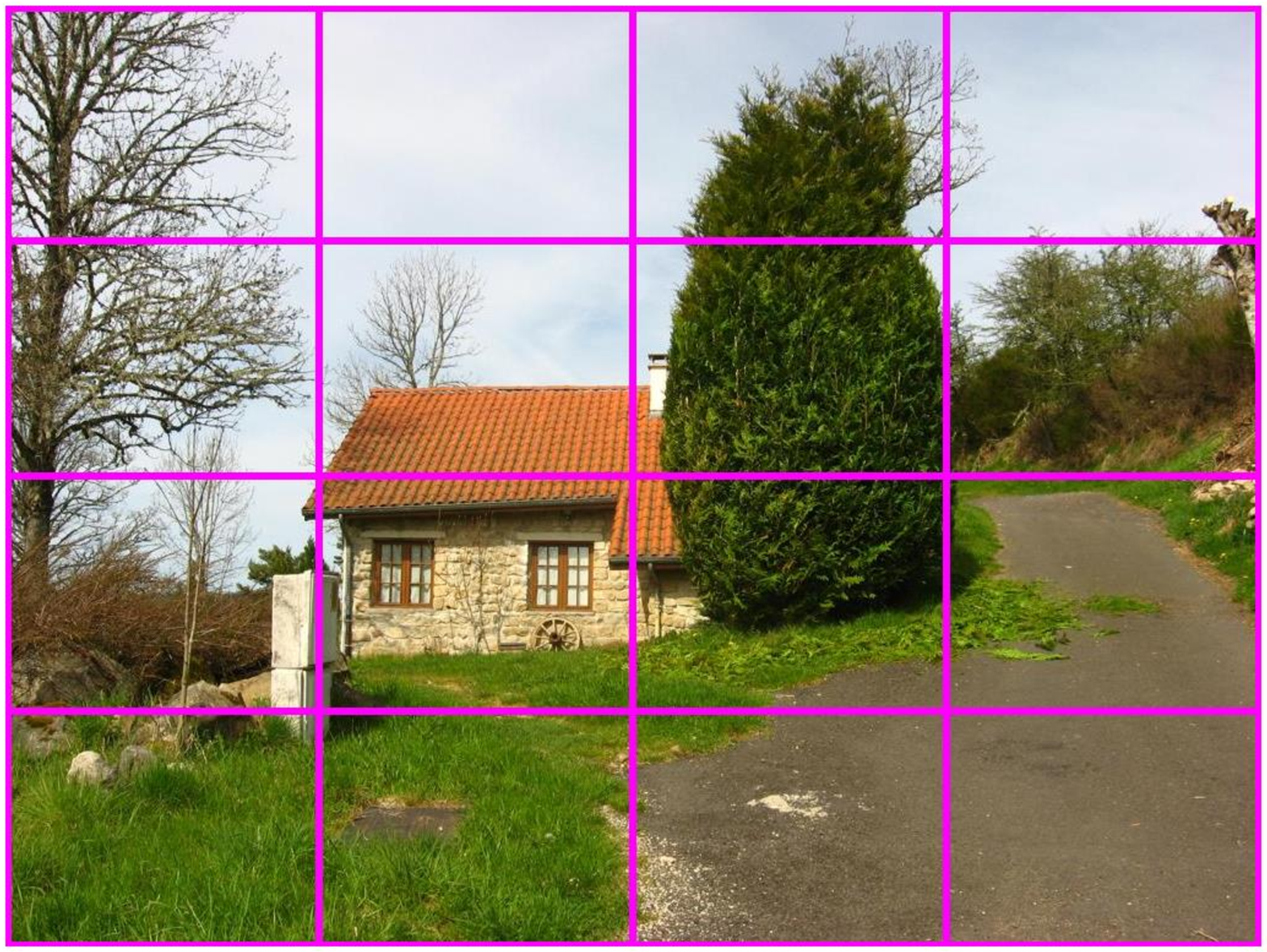}}
\hspace{0.2in}
 \subfigure[3rd scale]{\label{fig:level_three_partition}%
\includegraphics[width=2.1in]{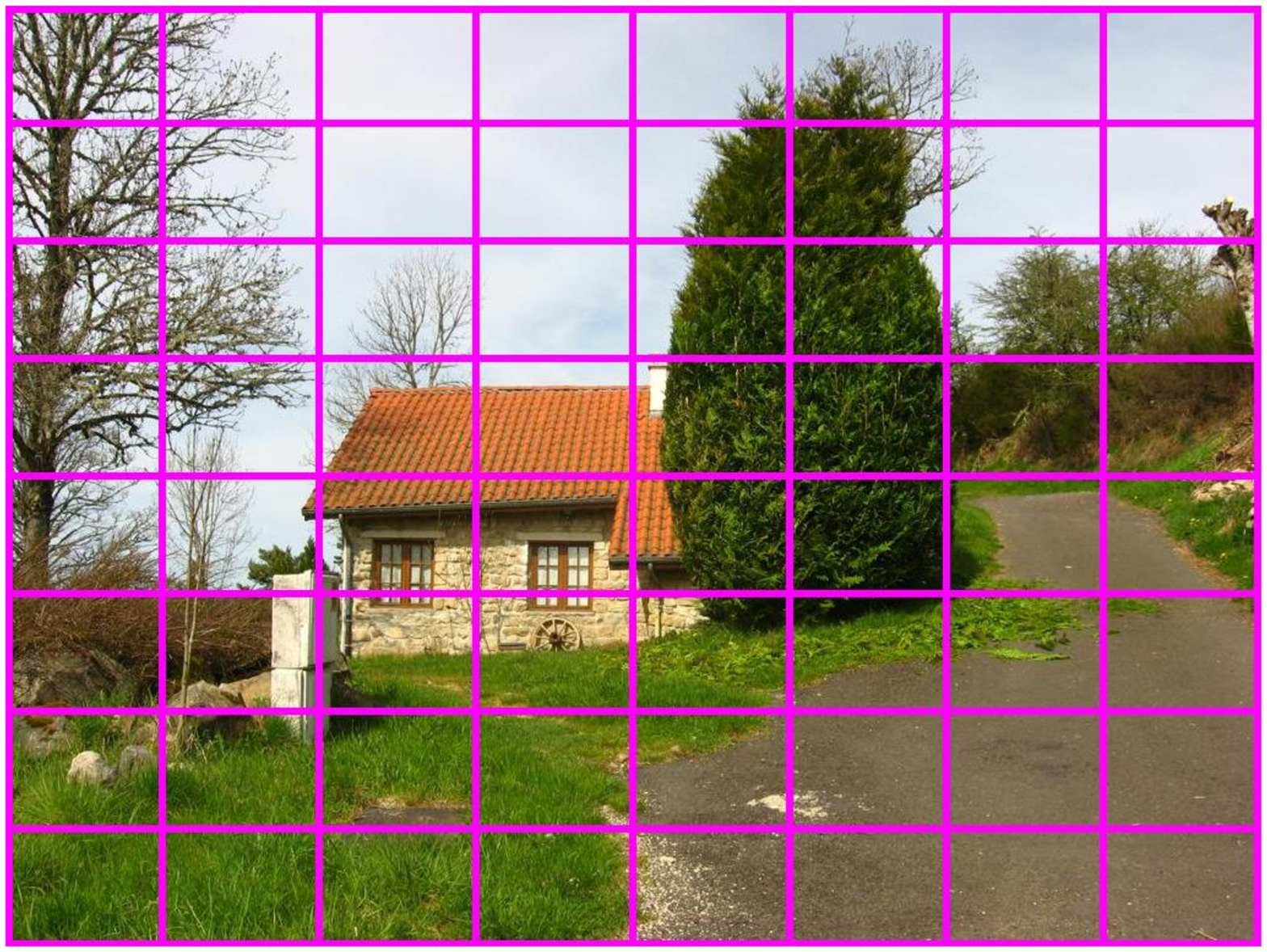}}
\caption{Strategy of image partitioning. Each image is partitioned with three scales. The first scale (a) considers the whole image. The second (b) and third (c) scales partition an image into 16 and 64 blocks, respectively. Correspondingly, the first scale characterizes the global context, while the last two scales describe regional context.}

\label{fig:partition strategy}
\end {figure*}

Features can be viewed as the engine of image retrieval.
Modern retrieval systems typically rely on local invariant features, such as SIFT, which shows satisfying performance. However, the SIFT feature is limited in many aspects. Specifically, it suffers from the lack of description of other attributes in an image, such as color, and it does not provide description of large image patches. Many strategies are proposed to provide complementary cues on different levels. On the local descriptor level, for example,
to augment the SIFT feature with color, Zheng \emph{et al.} \cite{zheng2014coupled} use the coupled multi-index structure, while the bag-of-color method \cite{BOC} embeds binary color signatures.
On the regional level, the bag-of-boundary approach \cite{arandjelovic2011smooth} partitions an image into regions. Similar with \cite{souvannavong2005region}, regions are described by multiple features. In another case, Fang \emph{et al.} \cite{fang2013giant} analyze geo-informative attributes in each region using a latent learning framework for location recognition. On the global level, features such as color histogram, spatial layouts, or attributes can be integrated with BoW using graph-based fusion \cite{zhang2012query, dengvisual}, co-indexing \cite{co_indexing}, or semantic hierarchies \cite{zhang2013attribute}. This paper departs from previous work by exploring the integration of all three levels of features using a principled probabilistic framework, providing theoretical insights on this topic.

Recently, deep learning models \cite{lecun1998gradient, hinton2006reducing} are receiving increasing attention. This class of learning machines works by constructing high-level features from low-level ones, thus automating the feature construction process with the feature hierarchy.
Among them, the convolutional neural network (CNN) \cite{lecun1998gradient} represents a deep learning model which applies trainable filters and pooling operations in an alternating manner. The resulting features are getting more complex and semantic aware along the hierarchy. When incorporated with appropriate regularization \cite{yu2008deep}, CNN has been shown to produce superior performance in various computer vision tasks, such as classification \cite{krizhevsky2012imagenet}, object detection \cite{girshick2013rich}, etc. Additionally, LeCun \emph{et al.} \cite{lecun2004learning} show that CNN also has some invariance to certain variations of the input image. In the field of image retrieval, the effectiveness of CNN  features has not been extensively studied, especially within the BoW model. In this paper, we make initial attempts on this issue, and provide feasible ways of integrating CNN features into the BoW structure.
\section{Feature Design}
\label{section:feature_design}
\subsection{Image Partitioning}
 \label{section:partition}
 In the framework of spatial pyramid matching (SPM) \cite{SPM}, features are extracted at a single scale and then pooled over increasing scales. Our work, instead, starts with a distinct idea: features are extracted at increasing scales. To this end, an image is partitioned into regions of three scales.

 Specifically, the first scale covers the whole image, corresponding to the global level context, as shown in Fig. \ref{fig:partition strategy}(a). The second and third scales (Fig. \ref{fig:partition strategy}(b) and Fig. \ref{fig:partition strategy}(c)) both encode regional context. For the second scale, each window is of size $h/4 \times w/4$, where $h$ and $w$ denote the height and width of the whole image, respectively.
 Similarly, the third scale is half the size of the second one: the window size is $h/8 \times w/8$.
 The second and third scales encode rotation invariance to some extent.

 Our partitioning strategy is simple. For each image, a fixed number of partitions are generated, \emph{i.e.,} $1+16+64 = 81$. The number of extracted CNN features per image is moderate, and it takes less than 2 seconds for feature extraction. Moreover, note that each keypoint is located within one global image, and two regions of different scales. We call the regional and global contexts as ``environment'' in this paper. 

\subsection{Feature extraction} In this paper, we extract a 4096-D feature vector from a partitioned region or the entire image. We use the pre-trained Decaf framework \cite{donahue2013decaf} which implements the convolutional neural networks with the goal of being efficient and flexible. Decaf takes as input an image patch of size $227 \times 227 \times 3$, with the mean subtracted. Features are calculated by forward propagation through five convolutional layers and two fully connected layers. Readers may refer to \cite{donahue2013decaf} for more details about Decaf. We will provide a comparison of features from the last two layers in Section \ref{section:global_feature}.

\subsection{Signed Root Normalization (SRN)}
\label{section:normalizatoin}
\begin{figure}[t]
\centering
\subfigure[Original Feature]{\label{fig:feature_origin}%
\includegraphics[width=3.1in]{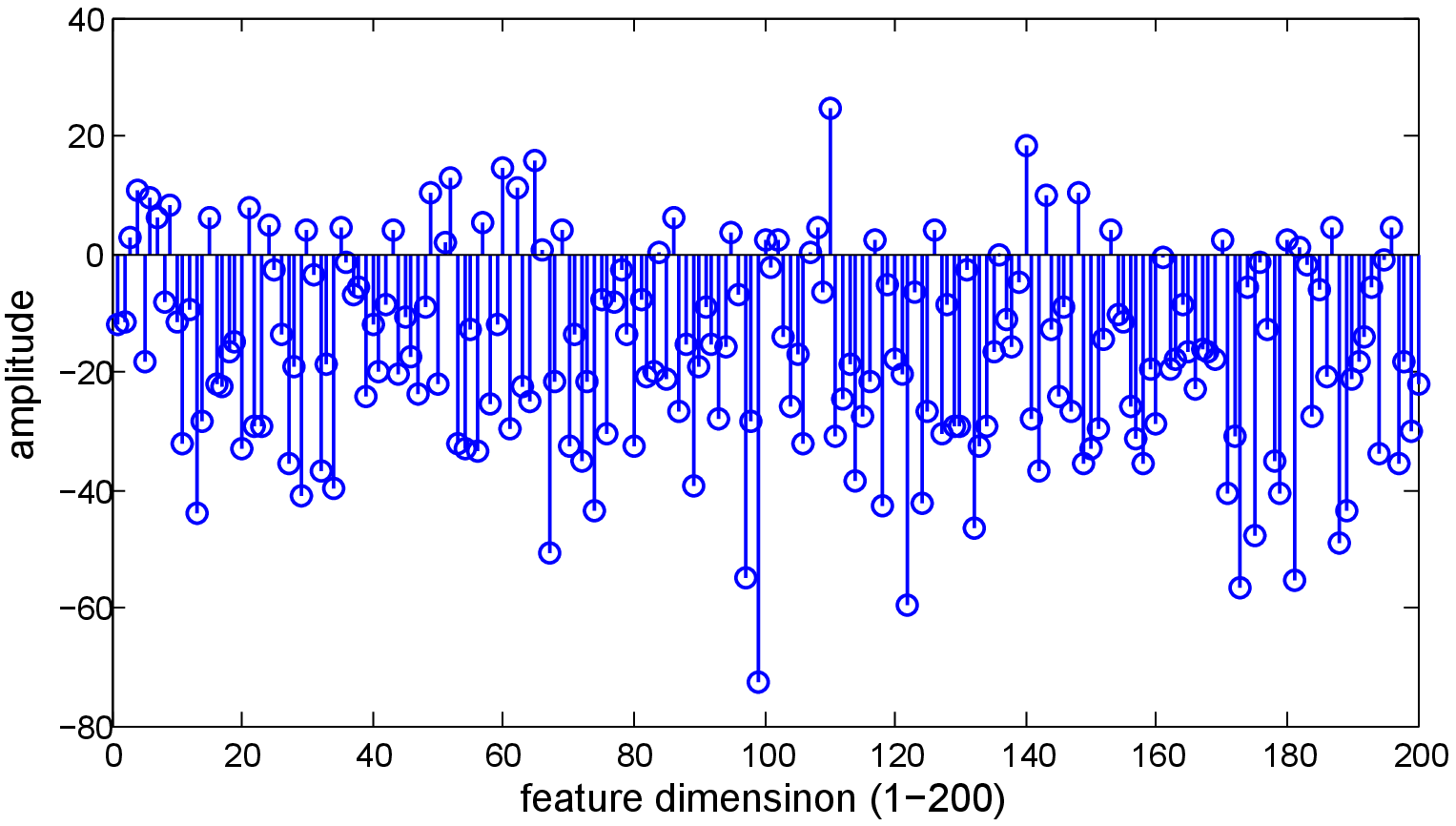}}
\subfigure[Normalized Feature]{\label{fig:feature_normed}%
\includegraphics[width=3.1in]{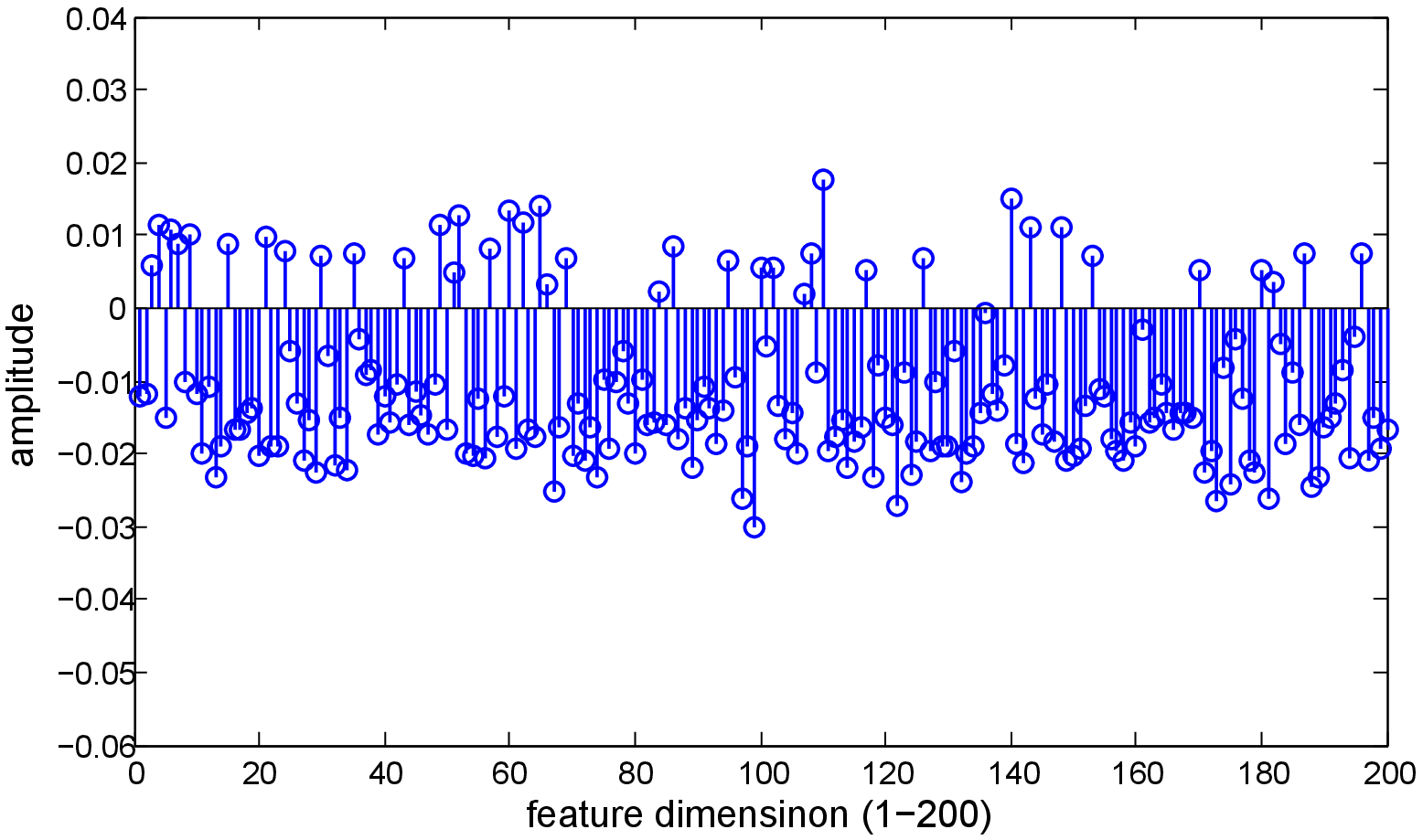}}
\caption{Illustration of Signed Root Normalization (SRN) on a CNN descriptor (dimension 1-200). The original (a) and normalized features are compared. We observe a more uniform data distribution after SRN.}
\label{fig:PQ_ET}
\end {figure}

The original CNN feature has a large variation in its value distribution: from -72.8 to 24.8 as the case shown in Fig. \ref{fig:feature_origin}. This is potentially problematic. For example, if two vectors differ a lot in one dimension, but similar in others, their Euclidean distance could be large. This problem would be more severe if we consider the fact that the negative values are produced by suppressed neurons, and convey less useful cues compared with the positive ones. If the difference occurs at negative value entry, the resulting distance is more unreliable. To address this problem, and thus produce more uniformly distributed data, we exert on each dimension the following function:
\begin{equation}\label{eq:normalization}
  f(x) = sign(x) |x|^{\alpha},
\end{equation}
where $sign(\cdot)$ denotes the signum function and $\alpha \in [0, 1]$ is the exponent parameter. Finally, the feature vector is $\ell_2$-normalized. In Section \ref{section:global_feature}, the parameter $\alpha$ is tuned and effectiveness of SRN will be demonstrated.

\makeatother

\subsection{Binary Signature Generation}
\label{section:binary_signature}
Given the high dimension of CNN vector and the requirement of memory efficiency, we transform the floating-point vector into a binary signature. In this step, we employ the well-known locality-sensitive hashing (LSH) \cite{charikar2002similarity}. Specifically, a hash key is obtained based on rounding the output of the product with a random hyperplane,
\begin{equation}\label{eq:LSH}
  h_r(x) = \begin{cases}
1, &\mbox{if $r^Tx \geq 0$} \\
0, &\mbox{otherwise}
\end{cases},
\end{equation}
where $r$ is a random hyperplane sampled from a zero-mean multi-variate Gaussian $\mathcal{N}(0, I)$ of the same dimension with $x$. For each CNN vector $x$, a total of $b$ hash keys are generated with $b$ hash functions by repeating Eq. \ref{eq:LSH} $b$ times. In our experiment, we set $b = 128$, thus producing a 128-bit binary signature for each CNN descriptor.

\section{Deep Embedding Framework}
\label{section:deep_embedding}
With the regional and global CNN features and the local SIFT feature, we introduce our feature fusion framework in this section.
\subsection{Model Formulation}
Given a query keypoint $x$ in image $q$ and an indexed keypoint $y$ in image $d$, we want to estimate the likelihood that y is a true match of $x$. In this paper, we define \emph{true match} as a pair of keypoints which are similar on local, regional, and global levels. This probability can be modeled as follows,
\begin{equation}\label{eq:model}
  f(x, y) = p(y \in T_x),
\end{equation}
where $T_x$ is the set of keypoints which are true matches to query keypoint $x$. To model the context cues, we define $\mathcal{E}_x$, $\mathcal{E}_y$ to be the ``environment'' variable of $x$ and $y$, respectively. Then, we have,
\begin{equation}\label{eq:environment_equal}
  p\left(\mathcal{E}_y = \mathcal{E}_x \right) + p\left(\mathcal{E}_y \neq \mathcal{E}_x\right) = 1,
\end{equation}
where $\mathcal{E}_y = \mathcal{E}_x$ means that $y$ belongs to $x$'s true match in terms of ``environment similarity'', and correspondingly $\mathcal{E}_y \neq \mathcal{E}_x$ indicates $y$ is a false environmental match. For simplicity, we denote $y\in T_x$ as $T_x$. Then, using Law of Total Probability, we get
\begin{equation}\label{eq:total_probability}
\begin{aligned}
  p(T_x) &= p(T_x \left.\right| \mathcal{E}_y = \mathcal{E}_x) \cdot p(\mathcal{E}_y = \mathcal{E}_x) \\
   &+ p(T_x \left.\right| \mathcal{E}_y \neq \mathcal{E}_x) \cdot p(\mathcal{E}_y \neq \mathcal{E}_x)
   \end{aligned}
\end{equation}
In the second term in Eq. \ref{eq:total_probability}, $p(T_x \left.\right| \mathcal{E}_y \neq \mathcal{E}_x)$, encodes the probability that $y$ is a true match of $x$ given that the contexts of $x$ and $y$ do not match. Clearly, according to our definition of \emph{true match}, this term equals to zero, \emph{i.e.,}
 \begin{equation}\label{eq:zero_term}
   p(T_x \left.\right| \mathcal{E}_y \neq \mathcal{E}_x) = 0.
 \end{equation}
 Therefore, we can readily neglect the second term.

Moreover, remember that the ``environment'' of a keypoint includes both global and regional contexts. Hence, $\mathcal{E}_x$ can be decomposed into $\mathcal{E}_x^g$ and $\mathcal{E}_x^r$, which represent global and regional contexts, respectively. Considering this, as well as the neglection of the second term, Eq. \ref{eq:total_probability} can be re-written as,
\begin{equation}\label{eq:model_final}
\begin{aligned}
  p(T_x) &= p(T_x \left.\right| \mathcal{E}_y = \mathcal{E}_x) \cdot p(\mathcal{E}_y = \mathcal{E}_x) \\
  &=p(T_x \left.\right| \mathcal{E}_y = \mathcal{E}_x) \cdot p(\mathcal{E}_y^r = \mathcal{E}_x^r, \mathcal{E}_y^g = \mathcal{E}_x^g)\\
  &=p(T_x \left.\right| \mathcal{E}_y = \mathcal{E}_x) \cdot p(\mathcal{E}_y^r = \mathcal{E}_x^r \left.\right| \mathcal{E}_y^g = \mathcal{E}_x^g) \cdot p(\mathcal{E}_y^g = \mathcal{E}_x^g).
  \end{aligned}
\end{equation}
In Eq, \ref{eq:model_final}, there involves three random variables to estimate, \emph{i.e.,} $p(T_x \left.\right| \mathcal{E}_y = \mathcal{E}_x) $ (\textbf{Term 1}), $p(\mathcal{E}_y^r = \mathcal{E}_x^r \left.\right| \mathcal{E}_y^g = \mathcal{E}_x^g)$ (\textbf{Term 2}), and $p(\mathcal{E}_y^g = \mathcal{E}_x^g)$ (\textbf{Term 3}). In Section \ref{section:probability_estimation}, the estimation of the three terms will be investigated.

\subsection{Probability Estimation}
\label{section:probability_estimation}
\noindent \textbf{Estimation of Term 1.} In Eq. \ref{eq:model_final}, Term 1 represents the likelihood of $y$ being a true match of query $x$ given that their ``environment'' matches on both global and regional levels. In other words, local features $x$ and $y$ belong to a pair of similar images, and are located in similar regions in the image. This situation is quite ideal: in terms of the definition of \emph{true match}, we only need to estimate the similarity on the local level, \emph{i.e.,} similarity between their local descriptors. Strictly speaking, the estimation requires labeled keypoints within such matched regions. But due to the lack of labeled data, we approximate this problem by relaxing the conditioned environment term and resorting to the classic estimation of local similarity. 

The baseline BoW model represents each local descriptor only by its visual word. Thus, the similarity between keypoints $x$ and $y$ is defined as,
\begin{equation}\label{eq: similarity_bow}
  s_b(x, y) = \delta_{q(x), q(y)},
\end{equation}
where $q(\cdot)$ is the quantization function maping a local feature to its nearest centroid in the codebook, and $\delta$ is the Kronecker delta response.

A good extension of BoW includes Hamming Embedding (HE) \cite{Hamming} that represents a local feature by both its visual word $q(x)$ and binary signature $b_x$. Given two features $x$ and $y$ quantized to the same visual word, their similarity function can be written as,
\begin{equation}\label{eq: similarity_he}
s^l(x, y) =
\begin{cases}
\exp\left(-d_H(b_x, b_y)^2 / \sigma^2\right),&h(b_x, b_y) < \kappa \\
0, &\mbox{otherwise}
\end{cases},
\end{equation}
where $d_H(\cdot)$ calculates the Hamming distance between $b_x$ and $b_y$, $\sigma$ is a weighting factor, and $\kappa$ is Hamming threshold. HE refines matching strength by considering the Hamming distance between features, thus improving retrieval accuracy. HE variants such as \cite{selective_match, qin2013query} design new weighting schemes, which do not deviate much from the basic idea. In this paper, we employ the original HE (Eq. \ref{eq: similarity_he}) as an estimation of Term 1.\\

\begin{figure}
  \centering
  \includegraphics[width=3.3in]{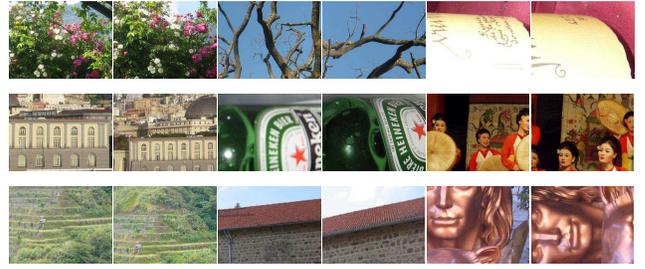}\\
  \caption{Examples of selected matching regions located in pairs of relevant images. The Euclidean distance between CNN features of these patches are calculated to estimate term 2.}
  \label{fig:patch_examples}
\end{figure}

\noindent \textbf{Estimation of Term 2.} Term 2 encodes the probability distribution of $\mathcal{E}_x^r$'s true matches given that the corresponding images are similar. In this paper, this distribution is modeled as a function of the Euclidean distance between similar regions located in similar images. To this end, images with ground truth are used.

\begin{figure}
  \centering
  \includegraphics[width=3.35in]{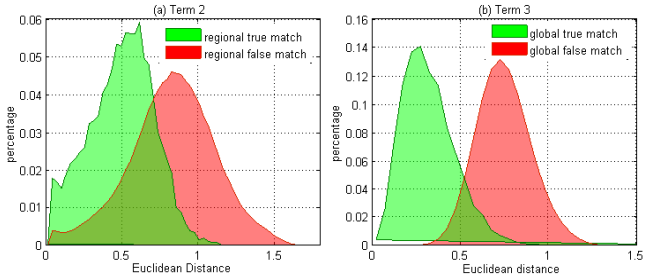}\\
  \caption{Euclidean distance distribution of regional (a) and global (b) matches. A clear separation can be observed.}\label{fig:distribution}
\end{figure}

\begin{figure}
  \centering
  \includegraphics[width=3.35in]{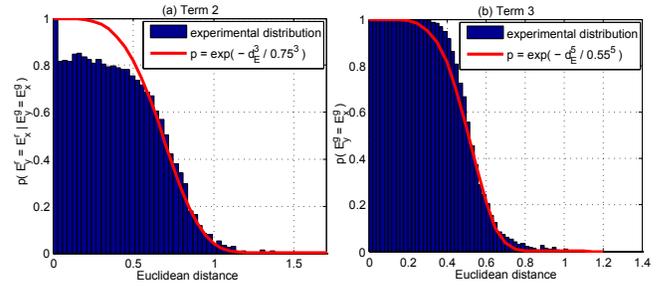}\\
  \caption{Probability distribution of Term 2 (a) and Term 3 (b). The general profile of the fitted curve (red) is used for testing.}\label{fig:polyfit}
\end{figure}

Specifically, we use the Holidays dataset \cite{Hamming} to perform empirical study. The statistical results are applied on other testing benchmarks. For this dataset, we DO know the ground truth which images are similar. But it DOES NOT provide ground truth which regions are visually true matches. Nevertheless, the fortunate thing is, the partition strategy (Section \ref{section:partition}) is simple, and generates a moderate number of regions per image. For this problem, we manually select visually similar regions from each pair of relevant images. Then, Euclidean distances between CNN features of these regions are computed, from which the distribution can be drawn. Note that, an image itself is also viewed as a relevant image and the data are collected in some pairs of identical images as well. Some examples of selected matching patches are shown in Fig. \ref{fig:patch_examples}. We plot the Euclidean distance distribution of regional true matches and false matches in Fig. \ref{fig:distribution}(a). The probability distribution of Term 2 is presented in Fig. \ref{fig:polyfit}(a).

From Fig. \ref{fig:distribution}(a), we can easily find that two distributions have a clear separation, with true regional matches on the left and false matches on the right. Therefore, we are able to softly distinguish the probability of a region being a true match to the query region. Figure \ref{fig:polyfit} demonstrates the feasibility of this argument: given the Euclidean distance between two regions, the matching strength can be determined automatically. We can approximate the  distribution in Fig. \ref{fig:polyfit}(a) with an exponential function,
 \begin{equation}\label{eq:similarity_region}
    s^r(x, y) = \exp\left(-d_E(c_x^r, c_y^r)^3 / \gamma^3\right),
 \end{equation}
 where $c_x^r$ and $c_y^r$ denote the regional CNN features for local points $x$ and $y$, respectively, $\gamma$ is a weighting parameter, and $d_E(\cdot)$ is the Euclidean distance. Here, we do not set a specific value to $\gamma$: we will tune the parameter in Section \ref{section:evalutation_deep_embedding}. We also notice that, the distribution is below the approximated curve when $d_E$ varies between 0 and 0.5. This is because some negative training samples may appear very similar, such as regions of grass, sky, etc. These scenes are very common in the images and may be randomly selected. According to the approximated function, we still assign a large likelihood to these matches, thus alleviating the impact of sampling error.
\\

\noindent \textbf{Estimation of Term 3.} Term 3 encodes the probability that  two images which contain $x$ and $y$ respectively are relevant ones. To measure this probability, global CNN feature is employed. Similar to the estimation process of Term 2, we plot the Euclidean distance distribution and the probability distribution in Fig. \ref{fig:distribution}(b) and Fig. \ref{fig:polyfit}(b), respectively. The profiles of these curves are similar to those of Term 2. Therefore, the similarity measurement can be written in a similar format. Assume that the global CNN vectors are $c_x^g$ and $c_y^g$, corresponding to two images, respectively. Their similarity, or $p(\mathcal{E}_y^g = \mathcal{E}_x^g)$, is defined as,
\begin{equation}\label{eq: similarity_global}
s^g(x, y) = \exp\left(-d_E(c_x^g, c_y^g)^5 / \theta^5\right),
\end{equation}
where $\theta$ is a weighting parameter.

\begin{figure*}
  \centering
  \includegraphics[width=7.0in]{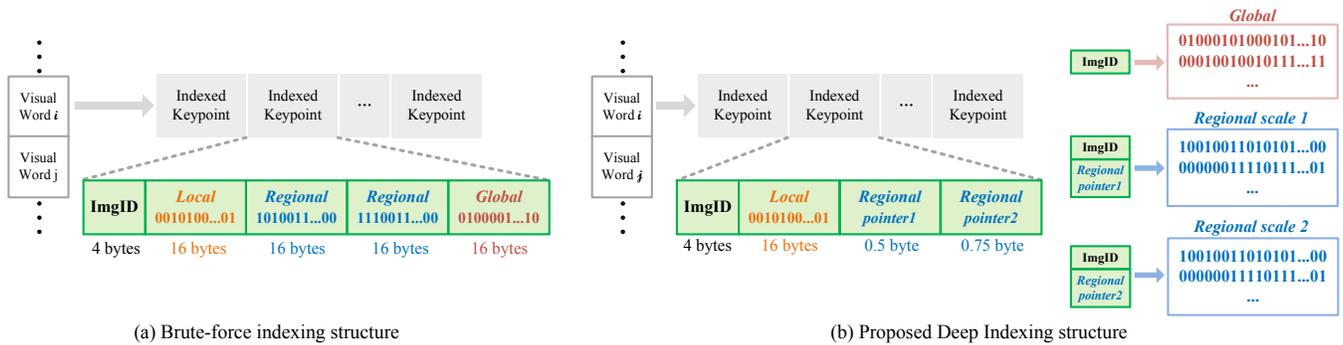}\\
  \caption{Inverted index organization for Deep Embedding. (a) The brute-force strategy embeds regional and global binary features with each indexed keypoint directly. (b) The proposed Deep Indexing structure stores regional and global features outside the inverted file. Each indexed keypoint stores instead two small pointers pointing to the regional features. The global features can be accessed via ImgID. Deep Indexing greatly reduces the memory usage.}\label{fig:inverted_index}
\end{figure*}

In the estimation of the Term 2 and Term 3, floating-point CNN vectors are used. The reason is that  full CNN vectors present a more precise data distribution, and the adoption of binarization serves as an approximation to the Euclidean space.In the experiments, we will present results obtained by both full and binarized vectors.

With the estimated probabilities, an explicit representation of the similarity model (Eq. \ref{eq:model} and Eq. \ref{eq:model_final}) can be provided: a combination of Eq. \ref{eq: similarity_he}, Eq. \ref{eq:similarity_region}, as well as Eq. \ref{eq: similarity_global}.

\subsection{Discussion}
\label{section:similarity_interpretation}
\noindent\textbf{Difficulties in feature embedding.}  Methods for global feature fusion varies from co-index \cite{co_indexing} to merging graphs of different rank results \cite{zhang2012query, dengvisual}. A common difficulty with this fusion task lies in choosing an effective weighting scheme, because local and global features may produce scores diverse in numerical values.

  The other issue is the fusion of local and regional features. For example, the Bag-of-Boundaries model \cite{arandjelovic2011smooth} concatenates various features in a region into a single vector, but this method may be sensitive to segmentation results. The Bag-of-Colors (BOC) \cite{BOC} and Coupled Multi-Index (c-MI) \cite{zheng2014coupled} methods employ the product of SIFT and local color similarities using their binary signatures. As we can see, the problem of feature fusion on multiple levels is not trivial. Many \emph{ad hoc} methods lack theoretical analysis and ``framarization''.\\

\noindent \textbf{Our interpretation.}
The motivation of our matching strategy is straightforward: Given two keypoints $x$ and $y$, their matching strength are jointly determined by three levels of contextual evidences, \emph{i.e.,} local, regional, and global.
We model this problem from the start of keypoint matching.
Apart from local feature which describes the keypoint itself, we further integrate the ``environment'' variable for co-description. The ``environment'' consists of regional and global contexts, which are implicitly induced by conditional probability formula. We can see that the final similarity (Eq. \ref{eq:model_final}) naturally forms a representation taking all three levels into account.

In our probabilistic model, the ``environment'' features could be replaced by other schemes. For example, when considering color features, previous work such as BOC, or c-MI can be well interpreted. In both cases, the color feature is treated as context of a keypoint, and the similarity measurement can be learnt in a similar way.  If we only consider local feature, Qin \emph{et al.} \cite{qin2013query} use a learning method very similar to ours to derive a more accurate Hamming embedding function than the original \cite{Hamming}. Therefore, the proposed method can serve as a principled framework of feature fusion on multiple levels. For completeness, in Section \ref{section:evalutation_deep_embedding}, we will replace CNN with color histogram to further validate our framework.

\subsection{Deep Indexing}
\label{section:indexing}
The inverted file is employed in most retrieval systems. In essence, each inverted list corresponds to a visual word in the codebook. Methods such as HE use a word-level inverted file, where the inverted list stores many ``indexed keypoints'' that are featured by the same visual word. An indexed keypoint contains related metadata, such as image ID, Hamming signature, etc.

Our method also uses a word-level inverted file. A brute-force indexing strategy is to store all three levels of binary signatures for an indexed keypoint, as illustrated in Fig. \ref{fig:inverted_index}(a). The drawback of this strategy is clear: the regional and global features do not have a one-to-one mapping with local keypoints, but in a one-to-many way. The strategy in Fig. \ref{fig:inverted_index}(a) thus consumes more memory than actually needed.

To reduce memory overload, we propose a Deep Indexing structure which is illustrated in Fig. \ref{fig:inverted_index}(b). For each indexed keypoint, its image ID and local binary signatures are left unchanged. For the regional features, we use two small pointers to encode their location in the image. For example, if two regional features of a keypoint are extracted from the 12th and 41th ones of the 4$\times$4 and 8$\times$8 windows, their regional pointers would be 12 and 41, respectively. Because the value of the regional pointers is no larger than $2^4=16$ and $2^6=64$, their memory usage is 0.5 byte and 0.75 byte, respectively. As with the global feature, it can be represented simply by image ID which is already indexed, so it does not require additional memory. During online query, the regional features can be accessed by a combination of image ID and their pointers, while global features are pointed by their image ID. In this manner, the Deep Indexing structure greatly reduces the memory usage.
\section{Experiments}
\label{section:experiments}
In this section, experimental results on three benchmark datasets will be summarized and discussed.
\subsection{Implementation and Experimental Setup}
\noindent \textbf{Features.} For the BoW baseline, we employ the method proposed by Philbin \emph{et al.} \cite{AKM}. For Holidays and Ukbench,  keypoints are detected by Hessian-Affine detector. For Oxford5k, the modified Hessian-Affine detector \cite{Paris} is applied, which uses gravity vector assumption to fix rotation uncertainty. Keypoints are locally described by the SIFT feature. The SIFT descriptor is further processed by $\ell_1$-normalization followed by component-wise square rooting \cite{root_sift}. The rootSIFT is shown to produce improved performance under Euclidean distance at no cost. \\

\noindent \textbf{Codebook.} Codebooks are trained by approximate k-means (AKM) \cite{AKM}. For Holidays and Ukbench, the training SIFT features are collected from the Flickr60k dataset \cite{Hamming}, while for Oxford5k, the codebook is trained on Paris6k dataset \cite{Paris}.
We use a codebook of size 65 $k$ for Oxford5k following \cite{selective_match}, and of size 20 $k$ for Holidays and Ukbench. \\

\noindent \textbf{Multiple assignment \& burstiness.} When multiple assignment (MA) is used, it is applied only on the query side to avoid memory overload. We empirically set MA = 3, so that three nearest neighbors are located. For a small codebook, the burstiness problem \cite{burstiness, zheng2013lp} is more severe. We combine our method with the intra-image solution \cite{burstiness} by square-rooting the TF of the indexed keypoints. We refer to the burstiness weighting as Burst in our experiments. For visual word weighting, we use the avgIDF proposed in \cite{zheng2013lp} instead of the classic IDF.  \\

\setlength{\tabcolsep}{3.5pt}
\begin{table}[t]
\centering
\caption{Details of the datasets used in the experiments.}
\begin{tabular}{|l||cccc|}
\hline
Datasets& \# images &  \# queries & \# descriptors &  Evaluation \\

\hline
\emph{Holidays}& 1491 & 500 & 4,455,091 & mAP \\
\hline
\emph{Oxford}& 5063 & 55 & 12,534,635 & mAP\\
\hline
\emph{Ukbench}& 10200 & 10200 & 19,415,079 & N-S score\\
\hline
\emph{MirFlickr1M} & 1,000,000& -&502,115,988 & -\\
\hline
\end{tabular}
\label{table:datasets}
\end{table}

\subsection{Datasets}
Our method is tested on three benchmark datasets, \emph{i.e.,} \textbf{Holidays \cite{Hamming}}, \textbf{Oxford5k \cite{AKM}}, and \textbf{Ukbench \cite{HKM}}. The Holidays dataset contains 1491 images, collected from personal holiday photos. 500 query images are annotated, most of which have 1-3 ground truth matches. Mean Average Precision (mAP) is employed to measure retrieval accuracy. The Oxford dataset consists of 5063 building images among which 55 are selected as queries. This dataset is challenging since its images undergo extensive variations in illumination, angle, scale, etc. mAP is again used for Oxford5k. The Ukbench dataset has 10200 images, manually grouped into 2550 sets, with 4 images per set. For each set, the images contain the same object or scene, and the 10200 images are taken as queries in turn. The accuracy is measured by N-S score, \emph{i.e.,} the number of relevant images in the top-4 ranked images. To test the scalability of our system, the \textbf{MirFlickr1M dataset \cite{huiskes08}} is added to the benchmarks. It contains 1 million images crawled from Flickr. A summarization of the four datasets is presented in Table \ref{table:datasets}.

\begin{figure}[t]
  \centering
  \includegraphics[width=2.8in]{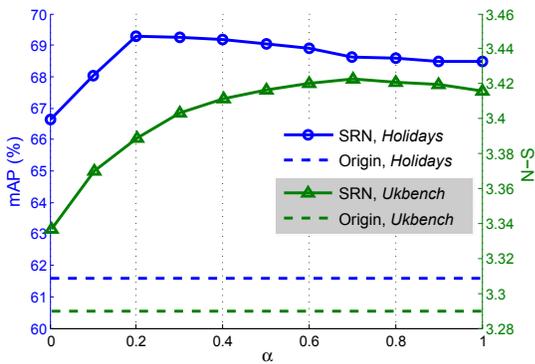}\\
  \caption{The impact of $\alpha$ on retrieval accuracy. Results on Holidays and Ukbench datasets are reported. }\label{fig:alpha}
\end{figure}

\makeatother
\begin{figure*}[t]
\centering
\subfigure[Tuning $\sigma$]{\label{fig:params_sigma}%
\includegraphics[width=2.2in]{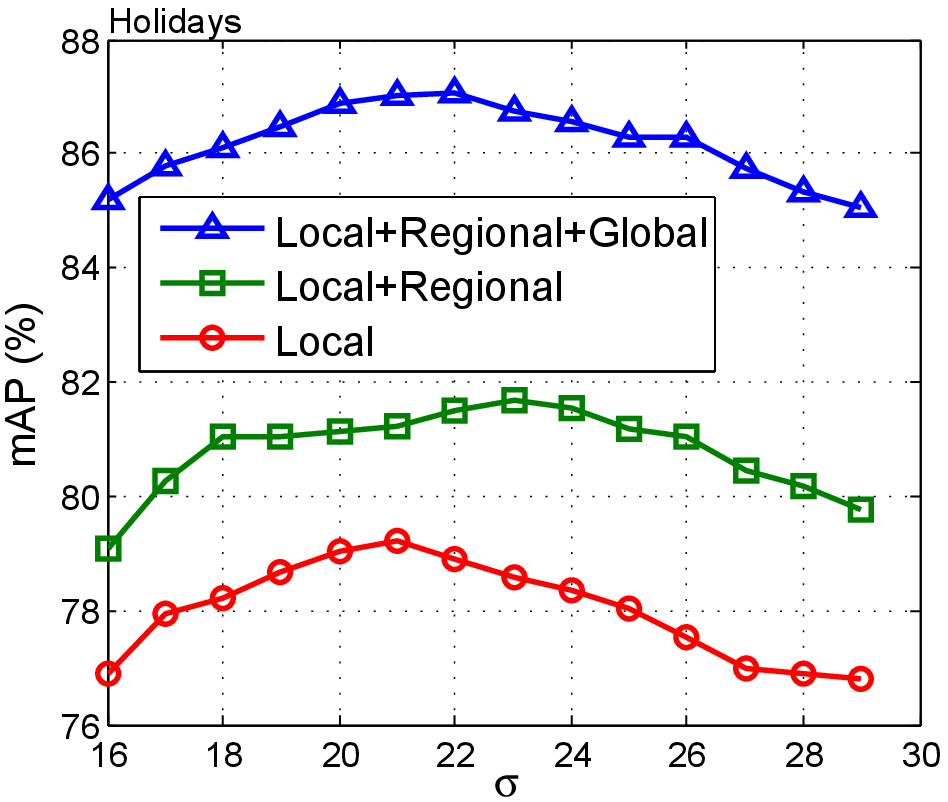}}
\hspace{0.1in}
\subfigure[Tuning $\gamma$]{\label{fig:params_gamma}%
\includegraphics[width=2.2in]{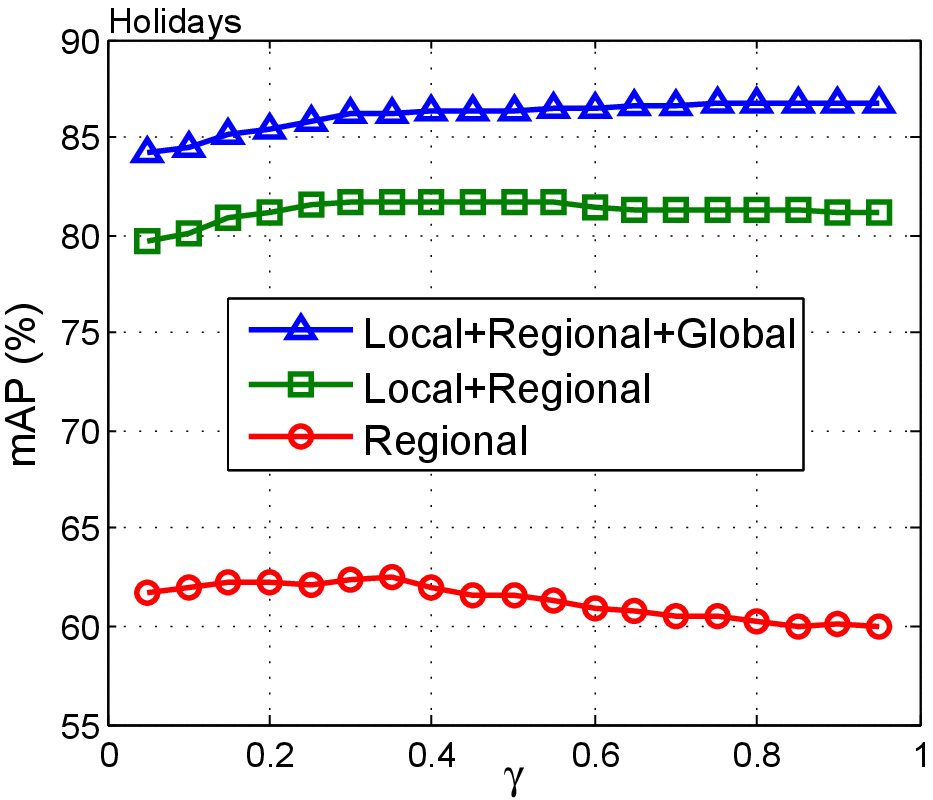}}
\hspace{0.1in}
\subfigure[Tuning $\theta$]{\label{fig:params_theta}%
\includegraphics[width=2.2in]{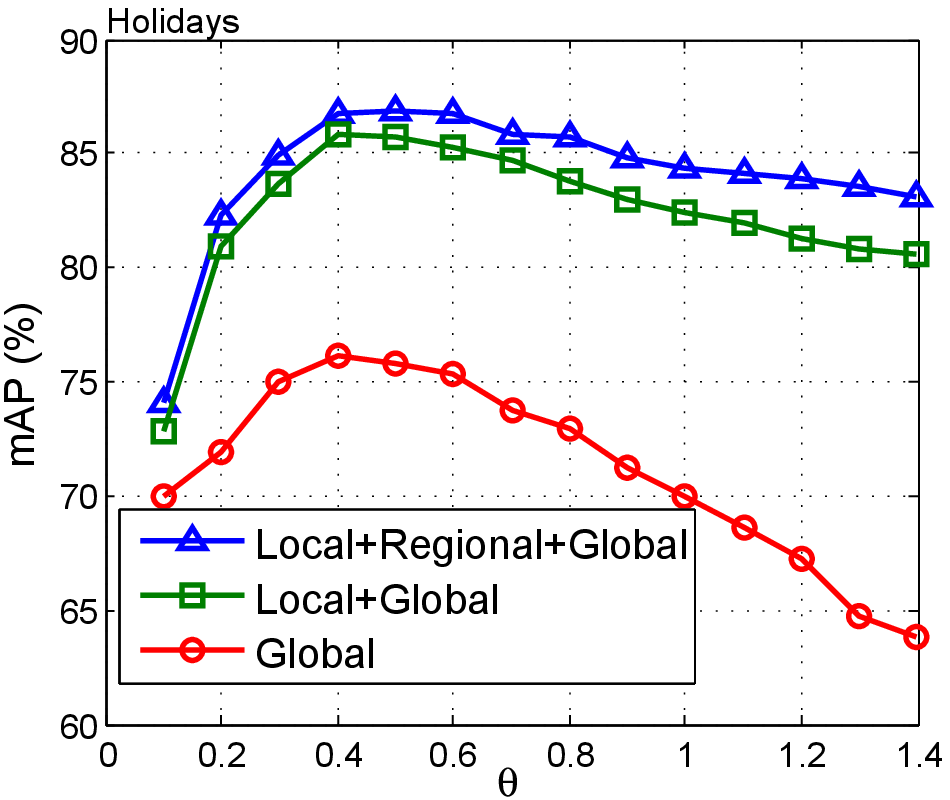}}
\caption{Parameter tuning process on Holidays dataset. Three parameters are considered, \emph{i.e.,} $\sigma$ (a) in Eq. \ref{eq: similarity_he}, $\gamma$ (b) in Eq. \ref{eq:similarity_region}, and $\theta$ (c) in Eq. \ref{eq: similarity_global}. For each parameter, different combinations of feature levels are shown.}
\label{fig:params_holidays}
\end {figure*}

\subsection{Global Feature Results}
\label{section:global_feature}
Using the normalization method described in Section \ref{section:normalizatoin}, we first test the impact of parameter $\alpha$ on the global feature performance. The results are shown in Fig. \ref{fig:alpha}.

We can see from Fig. \ref{fig:alpha} that the SRN method results in consistent improvements over the original feature in terms of linear search. For the Holidays dataset, original feature yields an mAP of 61.58\%, and the normalized feature produces mAP = 69.30\% at the peak. On Ukbench, similar observation can be derived: N-S score rises from 3.290 to 3.423 (best). From the two SRN curves, we set $\alpha$ to 0.5 considering the performance of both datasets. In the following experiments, $\alpha$ is kept.

We also compare the performance of CNN features of the two fully-connected layers (fc$_6$ and fc$_7$). The results are presented in Table \ref{table:different_layers}. ReLU (Rectified Linear Unit) stands for the operation in which the negative part of the CNN feature is cropped out (set to zero). We observe from Table \ref{table:different_layers} that ReLU has a marginal positive effect on Holidays, but is inferior on Ukbench and Oxford5k. Meanwhile, features from the sixth layer are superior to those from the seventh layer. We speculate that the fully connected layer may exert some negative effect on the features. Specifically, features extracted using ``fc$_6$'' obtains an N-S score of 3.416 on Ukbench, mAP of 69.03\% on Holidays, and 47.90\% on Oxford5k. Note that Oxford5k is actually an object retrieval dataset, so the global feature yields inferior performance to the BoW baseline. Although the ``fc$_6$-ReLU'' option produces a slightly higher result (69.13\%) on Holidays, we employ the features from the sixth layer without ReLU in the following experiments.

\subsection{Evaluation of Deep Embedding}
\label{section:evalutation_deep_embedding}
\noindent \textbf{Parameter Tuning.}
In the proposed framework, three major parameters are involved, \emph{i.e.,} weighting parameter $\sigma$, $\gamma$, and $\theta$ in Eq. \ref{eq: similarity_he}, Eq. \ref{eq:similarity_region}, and Eq. \ref{eq: similarity_global}, respectively. We vary the parameter values and report the performance on Holidays dataset in Fig. \ref{fig:params_holidays}.

\setlength{\tabcolsep}{7.0pt}
\begin{table}[t]
\caption{Retrieval accuracy on three datasets using CNN features from different layers.}
\centering
\begin{tabular}{|l||cccc|}
\hline
 Datasets & fc$_6$ & fc$_6$-ReLU & fc$_7$ & fc$_7$-ReLU \\

\hline
\emph{Ukbench}, N-S& 3.416 & 3.359 & 3.260 & 3.384 \\
\hline
\emph{Holidays}, mAP& 69.03 & 69.13 & 64.65 & 65.72 \\
\hline
\emph{Oxford5k}, mAP&47.90& 47.10& 39.35& 38.17\\
\hline
\end{tabular}

\label{table:different_layers}
\end{table}

\setlength{\tabcolsep}{9.5pt}
\begin{table*}[!t]
\caption{Image retrieval accuracy for three datasets by various methods.}
\begin{center}
\begin{tabular}{|l|c|c|c||cc|cc|cc|}
\hline
\multirow{2}{*}{Methods} &
\multirow{2}{*}{Local} &
\multirow{2}{*}{Regional} &
\multirow{2}{*}{Global} &
\multicolumn{2}{c|}{\emph{Ukbench}, N-S} &
\multicolumn{2}{c|}{\emph{Holidays}, mAP(\%)} &
\multicolumn{2}{c|}{\emph{Oxford5k}, mAP(\%)} \\

\cline{5-10}
&  &  &   & Deep*$^\text{1}$ & Deep & Deep* &Deep& Deep*& Deep\\
\hline
\hline

BoW &            &            &                        &3.109   & 3.109 & 50.10  &50.10& 53.01  &53.01\\
\hline
BoW & ${\times}$ &            &                        &3.560  & 3.560 &79.03  &79.03& 73.67 &73.67\\
\hline
BoW &            & ${\times}$ &                        & 3.441  &  3.315 & 65.60   & 62.37& 60.84 & 57.54\\
\hline
BoW &            &            & ${\times}$             & 3.706  & 3.589 & 76.11    &72.43 &  58.11 & 55.33\\
\hline
BoW & ${\times}$ &            & ${\times}$             & 3.762  & 3.668 & 85.60    &  83.47 &  75.57 & 74.21\\
\hline
BoW & ${\times}$ & ${\times}$ &                        &  3.681 & 3.629   &82.49     & 81.69  & 77.95 & 75.69\\
\hline
BoW & ${\times}$ & ${\times}$ & ${\times}$             & 3.832 & 3.751 &86.81& 84.20& 79.89 & 76.97\\
\hline
+ MA + Burst & ${\times}$ & ${\times}$ & ${\times}$    & \bf 3.851 & 3.790 & \bf88.08 &85.30& \bf 82.00& 80.02\\
\hline
+ post-processing &${\times}$&${\times}$&${\times}$&\emph{3.873}&3.844&\emph{87.32}&86.14&\emph{87.21}& 85.35\\
\hline
\end{tabular}
\begin{tablenotes}
\item$^\text{1}$ Throughout the paper, * denotes the case where floating-point CNN vector is used. Otherwise, binary CNN feature is referred to.
\end{tablenotes}
\label{table:various_approaches}
\end{center}
\end{table*}
\setlength{\tabcolsep}{1.4pt}

\makeatother
\begin{figure*}[t]
\centering
\subfigure[Holidays]{\label{fig:color_holidays}%
\includegraphics[width=2.2in]{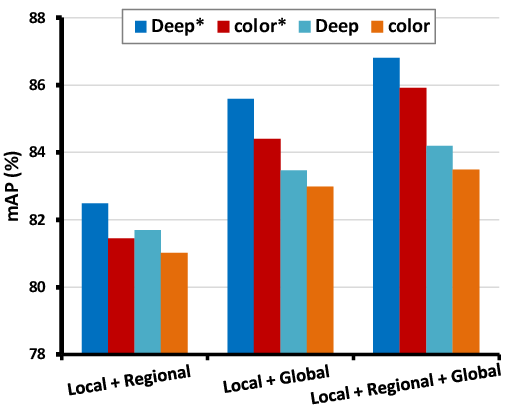}}
\hspace{0.1in}
\subfigure[Oxford5k]{\label{fig:color_oxford}%
\includegraphics[width=2.2in]{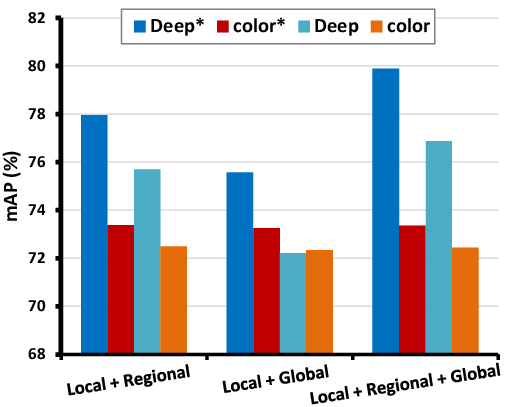}}
\hspace{0.1in}
\subfigure[Ukbench]{\label{fig:color_ukbench}%
\includegraphics[width=2.2in]{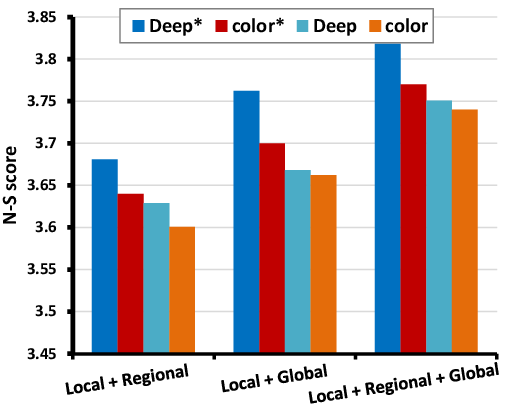}}
\caption{Comparison of HS histogram with CNN features. For each image patch, a 1000-D HS histogram is extracted and replace CNN features in the fusion framework. * means that floating-point vectors are used, while in other cases, features are binarized.}
\label{fig:color}
\end {figure*}

Overall, the profile of most curves in Fig. \ref{fig:params_holidays} first rises and then drops as the corresponding parameter increases. This is technically sound because the exponential curves shown in Fig. \ref{fig:polyfit} also have a transition point where the second derivative equals zero. From these results, we set $\sigma = 21$, $\gamma = 0.8$, and $\theta = 0.4$ in the following experiments. Note that the values of $\gamma$ and $\theta$ are close to those in Fig. \ref{fig:polyfit}. As with the Hamming threshold $\kappa$ in Eq. \ref{eq: similarity_he}, as previous work \cite{zheng2014coupled, Hamming} suggests, it produces steady performance when being set to a relatively large value. In this paper, we set $\kappa$ to 60. \\

\noindent \textbf{Contribution of the three levels.} In our method, visual matching is checked on local, regional, and global levels. Here, we analyze the contribution of each part as well as their combinations.

Fig. \ref{fig:params_holidays} demonstrates the trends of different level combinations on Holidays dataset. We can observe that under each parameter setting, the joint effect of multiple level evidences always brings benefit over single levels. Specifically, as indicated in Table \ref{table:various_approaches}, when used alone, the three levels of features produce mAP of 79.03\%, 65.60\%, and 76.11\% on Holidays, respectively. The integration of regional or global features with local HE obtains an mAP of 82.49\% (+3.46\%) or 85.60\% (+6.57\%), respectively. When three levels of evidences are jointly employed in the proposed framework, compared with the BoW baseline, we obtain N-S = 3.832 (+0.723), mAP = 86.81\% (+36.71\%), and mAP = 79.89\% (+26.88\%) on the Ukbench, Holidays, and Oxford5k datasets, respectively. These results strongly prove that the contextual cues of CNN features are perfectly complementary to the local features.

Moreover, we also find that the regional features somewhat have less positive impact than global features on Holidays and Ukbench, but work better on Oxford5k instead. For example, on the Ukbench dataset, when local cues are not combined, the regional and global CNN descriptor alone yields improvements of +0.332 and +0.597 in N-S score over the baseline, respectively. On Oxford5k, the improvements are +7.83\% and +5.10\% in mAP, respectively. The reason is that, images in the Oxford5k dataset vary intensively in illumination and view changes, while images in the Holidays and Ukbench datasets are more consistent in appearance. Therefore, the global CNN descriptor is less effective on Oxford5k than on Ukbench and Holidays. The above conclusions can also be drawn from the binarized CNN features. \\

\noindent\textbf{Comparison of CNN and color histogram.} After showing that the framework is effective in incorporating contextual evidences, we seek to evaluate the effectiveness of CNN features in its descriptive power. To this end, we compare the results obtained by CNN feature and HS histogram on the three datasets. Specifically, a 1000-D HS histogram is extracted from each image patch in place of CNN. Following \cite{zheng2014coupled}, a component-wise square root is exerted on the HS histogram, followed by an $\ell_2$-normlizaiton. The results are presented in Fig. \ref{fig:color}.

We can clearly see that for all methods, \emph{i.e.,} ``Local + Regional'', ``Local + Global'', and ``Local + Regional + Global'', the CNN feature outperforms the color histogram. This can be attributed to the fact that CNN describes both texture and color features which is determined by its training process. This property brings additional descriptive power which single color feature lacks. The advantage is more obvious on Oxford5k dataset, where color feature loses its power due to the large illumination changes.\\

\begin{figure}
  \centering
  \includegraphics[width=3.35in]{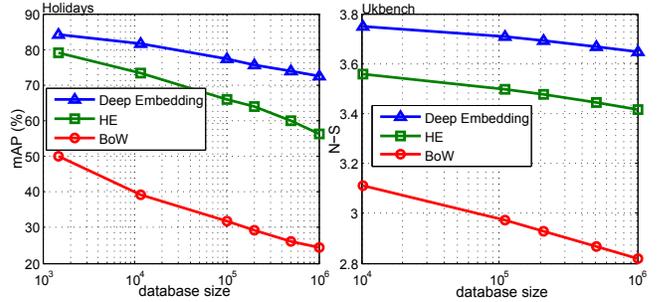}\\
  \caption{Large-scale experiments on Holidays and Ukbench datasets.}\label{fig:large_scale}
\end{figure}

\setlength{\tabcolsep}{4pt}
\begin{table}[t]
\centering
\caption{Memory cost of Deep Embedding.}
\begin{tabular}{|l|ccc|}
\hline
\multirow{2}{*}{Components}&
Per feature  &
Per image$^\text{1}$   &
1M dataset  \\

& (bytes)& (bytes)& (GB)\\
\hline
\hline
ImgID  &  4   & 4 $\times$ 500 & 1.87\\
TF & 1      &  1 $\times$ 500 & 0.47 \\
Local  & 16 & 16 $\times$ 500 & 7.48\\
Regional & 0.5+0.75 & 16 $\times$ 80 + 1.25 $\times$ 500 & 1.78\\
Global & 0 &  16 & 0.01\\
\hline
Total & 22.25 & 12.13 KB & 11.61\\
\hline
\end{tabular}

\begin{tablenotes}
\item$^\text{1}$ We assume that an image has 500 keypoints on average.
\end{tablenotes}
\label{table:memory}
\end{table}

\setlength{\tabcolsep}{4.7pt}
\begin{table}[t]
\caption{Memory cost and query time for different approaches on Holidays + MirFlickr1M dataset.}
\renewcommand{\arraystretch}{1.2}
\centering

\begin{tabular}{|l|cccc|}
\hline
Methods& \multirow{1}{*}{BoW} & 64-bit HE & 128-bit HE & Ours \\

\hline
\hline

Memory Cost (GB) & 1.87      &  5.61  &  9.35  & 11.61\\
\hline
Query Time (s)& 2.70& 2.11 & 2.13 & 2.32\\
\hline
\end{tabular}

\label{table:memory_comparison}
\end{table}

\setlength{\tabcolsep}{10pt}
\begin{table*}[t]
\caption{Performance comparison with state-of-the-art methods without post-processing}
\centering
\begin{tabular}{|l|cc|ccccccccc |}
\hline
Methods& Deep$^*$ &Deep& \cite{zheng2014coupled} &  \cite{selective_match}$^*$ & \cite{selective_match} & \cite{shen2012object} & \cite{co_indexing} & \cite{BOC} & \cite{jegou2010improving} & \cite{qin2013query}  & \cite{burstiness}\\

\hline
\hline
\emph{Ukbench}, N-S score& \bf 3.85  &3.79& 3.71   & - & - &  3.52  & 3.60 & 3.50  & 3.42 & - & 3.54\\
\hline
\emph{Holidays}, mAP(\%)& \bf 88.1 &85.3&84.0   & 82.2 & 81.0& 76.2  & 80.9  & 78.9 & 81.3 & 82.1 & 83.9\\
\hline
\emph{Oxford5k}, mAP(\%)& \bf 82.0 &80.0 &-& 81.7 & 80.4&75.2&68.7&-&61.5&78.0&64.7\\
\hline
\end{tabular}
\label{table:state_of_art_no_post}
\end{table*}

\noindent \textbf{Large-scale experiments.} To test the scalability of the proposed method, we perform large-scale experiments by populating Holidays and Ukbench with the MirFlickr1M dataset. We plot  accuracy against different database sizes, as shown in Fig. \ref{fig:large_scale}.

From Fig. \ref{fig:large_scale}, we can see that our method yields consistently higher performance over both the baseline and HE methods. Moreover, as the database gets scaled up, the performance gap is getting larger too. For example, on Holidays + 1M dataset, our method achieves an mAP of 72.4\%, while BoW and HE obtain 24.3\% and 56.3\%, respectively. 

The memory cost of the Deep Embedding method is calculated in Table \ref{table:memory}. For each indexed keypoint, 4 bytes are allocated for its Image ID (ImgID). Since we use the burstiness weighting strategy, one byte is consumed to store the TF data. Then, the local binary signature uses 16 bytes (128 bits). The two regional pointers takes 1.25 byte memory as analyzed in Section \ref{section:indexing}. On the other hand, on the image level, since each image is partitioned into 81 blocks (80 regional and 1 global), the memory cost for the binary CNN features is 16$\times$81 bytes. Hence, for the MirFlickr1M dataset, the total memory consumption arrives at 11.61 GB.

\setlength{\tabcolsep}{10pt}
\begin{table*}[t]
\caption{Performance comparison with state-of-the-art methods with post-processing}
\centering
\begin{tabular}{|l|cc|ccccccccc|}
\hline
Methods& Deep*&Deep&\cite{zheng2014coupled} & \cite{dengvisual}&  \cite{zhang2012query} &  \cite{qin2011hello} & \cite{shen2012object} & \cite{jegou2010improving} & \cite{qin2013query}  & \cite{burstiness} & \cite{root_sift}\\

\hline
\hline
\emph{Ukbench}, N-S score& \bf 3.87& 3.84 & 3.85& 3.75& 3.77 & 3.67 & 3.56   & 3.55 & - & 3.64 &-\\
\hline
\emph{Holidays}, mAP(\%)& \bf 87.3 & 86.1 & 85.8 &84.7& 84.6 & -   &  -   & 84.8 & 80.1 &  84.8 &-\\
\hline
\emph{Oxford5k}, mAP(\%)& 87.2 & 85.4 &-& 84.3&-& 81.4& \bf 88.4& 74.7&85.0& 68.5 &80.9\\
\hline
\end{tabular}

\label{table:state_of_art_post}
\end{table*}

We also compare our memory usage with the baseline and HE methods in Table \ref{table:memory_comparison}. The BoW baseline, also the one presented as ``ImgID'' in Table \ref{table:memory}, uses 1.87GB memory. HE is the sum of ``ImgID'' and ``Local'' in Table \ref{table:memory}. The 128-bit HE consumes a memory of 9.35 GB. The proposed Deep Embedding method exceeds 128-bit HE by 2.26 GB. This difference major consists of the storage of TF (which can be discarded since Burstiness does not bring much improvement) and the regional binary signatures.

\begin{figure*}
  \centering
  \includegraphics[width=6.9in]{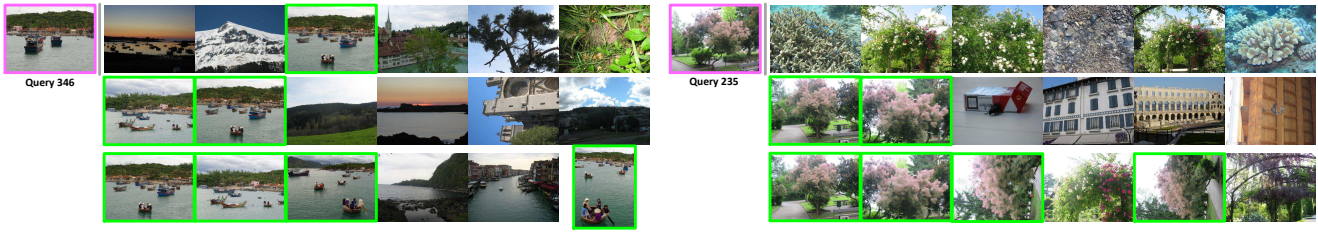}\\
  \caption{Sample retrieval results on Holidays dataset. The query image is on the left. Three methods are compared, \emph{i.e.,} BoW (first row), HE (second row), and Deep Embedding (third row).}\label{fig:sample_results}
\end{figure*}

On the other hand, Table \ref{table:memory_comparison} also compares the query time for the three methods. On the Holidays + 1M dataset, it takes 2.70s on average for a query, while our method consumes 2.32s. Compared with HE, since our method involves more Hamming distance calculation, the query time is marginally longer. The above analysis demonstrates that our method only marginally increases memory usage and query time over HE, but brings much higher retrieval accuracy, thus proving its effectiveness in large-scale settings.\\

\noindent\textbf{Comparison with state-of-the-arts} We compare our results with state-of-the-art methods in Table \ref{table:state_of_art_no_post} and Table \ref{table:state_of_art_post}. First, when no post-processing steps are involved, the proposed Deep Embedding yields superior performance, as shown in Table. \ref{table:state_of_art_no_post}. Notably, we achieve \textbf{N-S = 3.85} on Ukbench, \textbf{mAP = 88.1\%} on Holidays, and \textbf{mAP = 82.0\%} on Oxford5k, respectively. Note that, for the comparison with Tolias \emph{et al.} \cite{selective_match}, we compare with the reported results obtained with the same number of SIFT descriptors. We speculate that when more local features are extracted for each image, our method will bring more benefit because a higher recall is provided.

 Reranking steps are effective in boosting multimedia retrieval performance \cite{natsev2007semantic, pang2013ranking}. In our work, for Ukbench and Holidays datasets, Graph Fusion \cite{zhang2012query} with global CNN feature is employed; for Oxford5k, we use Query Expansion \cite{query_expansion} on the top-ranked 200 images. In Table \ref{table:state_of_art_post}, we achieve \textbf{N-S = 3.87}, \textbf{mAP = 87.3}, and \textbf{mAP = 87.2} on the three datasets, respectively, which are very competitive with the state-of-the-arts. We notice that on Oxford5k, our result is slightly lower than \cite{shen2012object} with reranking, but much higher than \cite{shen2012object} without reranking. This is probably because \cite{shen2012object} uses a more sophisticated reranking method,

In Fig. \ref{fig:sample_results}, two groups of visual ranking results are provided. Since the CNN feature is trained on labeled data, semantic cues can be preserved, and our method (the third row) is able to return challenging candidates which are semantically related to the query.

\section{Conclusions}
\label{section:conclusions}
In this paper, the Deep Embedding framework is proposed. Our motivation is two-fold. First, when matching pairs of keypoints, contextual evidences should be integrated with local cues, namely, the regional and global descriptions. Second, the successful CNN activation feature is rarely used in BoW based image retrieval. Here we employ CNN features to describe regional and global patches, which provides a feasible solution to CNN usage. Our method is built on the probabilistic derivation of a feature being true match of a given query. We show that our model successfully integrates multiple levels of contextual cues, which greatly reduces the impact of false positive matches.

Extensive experiments on three benchmark datasets show that the three levels of features are well complementary and improve significantly over the baseline. When combined in the Deep Embedding framework, we are capable of producing superior performance to the state-of-the-art methods.

This paper demonstrates the effectiveness of CNN feature in the image retrieval as auxiliary cues to the classic BoW model. A possible future direction is to use CNN feature as the major component, and build effective and efficient patch-based retrieval systems.

{\footnotesize
\bibliographystyle{abbrv}
\bibliography{sigproc} 
} 

\end{document}